# A ROBUST BI-DIRECTIONAL ALGORITHM FOR PEOPLE COUNT IN CROWDED AREAS

Dr.P.Satyanarayana[1], P.Gopikrishna[2], K.Soukhya[3], M.Satvik[4], Y.CharanKumar[5]
Professor[1], Department of ECE, K L University, Guntur, India
satece@kluniversity.in[1]
U G Students[2,3,4,5], Department of ECE, K L University
gopikrishnapavuluri23@gmail.com[2],
soukhya.kunda@gmail.com[3], satvik.mente@gmail.com[4,5]

**Abstract:** People counting system in crowded places has become a very useful practical application that can be accomplished in various ways which include many traditional methods using sensors. Examining the case of real time scenarios, the algorithm espoused should be steadfast and accurate. People counting algorithm presented in this paper, is centered on blob assessment, devoted to yield the count of the people through a path along with the direction of traversal. The system depicted is often ensconced at the entrance of a building so that the unmitigated frequency of visitors can be recorded. The core premise of this work is to extricate count of people inflow and outflow pertaining to a particular area. The tot-up achieved can be exploited for purpose of statistics in the circumstances of any calamity occurrence in that zone. Relying upon the count totaled, the population in that vicinity can be assimilated in order to take on relevant measures to rescue the people.

**Keywords:** counting system, blob, inflow, outflow

## I. Introduction

There exist many works related to the people count, which includes many traditional methods using sensors. The people detection techniques based on various sensors like Infrared beam sensors, thermal based sensors, foot step sensors which involved pressure sensing methods[1]. The conventional methods were not be accurate and reliable.

The traditional people detection methods cannot identify whether the moving object is human. The conventional systems using sensors have a chance of miscounting the people. The systems might assume the vehicles or any animal to be a human as they are also dynamic leading to the wrong count and the sensor based people counting systems can be used in the small crowded areas but cannot be used for large crowded areas[2]. The people counting systems using sensors are not reliable and cannot be preferable for the real time situations and so the image processing algorithms had developed in order to improve the efficiency. In the path of the image processing, a people counting system using background subtraction was developed which became a vital part in the process of people detection[3]. Counting systems based on computer-vision provides an effective path for this work. The foremost thing to be done in the computer-vision systems is to segregate the moving object (in this case people) from the static background scene. Through which the counting can be done followed by the process of the foreground extraction. Computer vision algorithms were developed in order to implement the people counting system which can give better results compared to the existing algorithms[4]. There were different algorithms for counting the people using various classifiers like Haar cascade classifier[5], which involves usage of much memory for the trained data set[6]. Neural network based algorithms were developed to improve the accuracy obtained through existing algorithms[7]. A comparison was made in [4] among the various existing algorithms like background subtraction, Hough circle transform and Histogram of Oriented Gradients (HOG). One of the methods of people detection was the use of the virtual lines for the people count[8] that involves the counting of the people when the virtual line in the field of view of camera had been crossed by any obstruction. This had a drawback which was because of the usage of the single virtual line that increments the counter when a person crosses the virtual line. In case of people flow after one by one, this method can be accurate. But, in case of the people entering or leaving the place at the same time, ambiguity occurs and the system fails in the proper count of the people. Another method called component based human detection for counting the people in a crowd can be adopted by the extraction of features from the processed frames. There are many features that can be extracted from the background subtracted frame that can be matched to the human features like frontal face, shoulders, upper body, lower body, facial features[9] and so on leading to much usage of memory for the human detection, which detects the moving object as a human.

It is important to know the direction of the people along with the count in order to make the implemented algorithm efficient. The proposed people counting algorithm corresponding to this paper using blob analysis provides the count of the people entering and leaving a particular area along with the direction of the moving people which in addition overcomes the problem of occlusion in the case of more number of people move in the cameras field of view.

## II. Literature Survey

There exists many works related to the people count, which has many traditional methods based on the image sensing techniques .The people detection based techniques using various sensors like Infrared beam sensors, thermal based sensors, foot step sensors which involves pressure sensing methods and so on. The conventional methods include the usage of turnstiles, which may not be accurate and also makes the people feel difficult to go through that particular way. The traditional people detection methods cannot identify whether the moving object is human. The conventional systems using sensors have a chance of miscounting the people. The systems might assume the vehicles or any animal to be a human as they are also dynamic leading to the wrong count.

Consider, the people detection using IR beam sensors which considers any object that crosses the beam as a count. The limitation in IR based counters was that any object which crosses the infrared beam is counted as a person leading to miscount. There exists a source and a receiver in the system. The beam emitted by the source will reach the receiver if there exist any obstacle and get absorbed if there is no obstruction. In case of any object moving in between the wall and the source, the beam gets obstructed and reflects to the receiver placed below the source. The receiver then gets a voltage if the beam gets reflected resulting in the increment of the count. This can be done with the help of a micro-controller. This system also has its limitation which is due the miscount that might happen due to any obstacle other than human.

Like the IR beam counter, the same principle was followed in the case of laser beam counter in which there was a chance of counting only a single crossing at a time. If two or more persons cross the beam at the same time then the count can be incremented only by one which leaded to corrupted result. Those conventional systems can be suitable only for a narrow entrances but not for wide entrances. In order to avoid all these failures in the system, the preferable method for counting the people can be using the image processing algorithms which gives the efficient results and are reliable.

Another conventional counter was using the thermal sensors. The people counter based on using the thermal sensors depend on the temperature of the objects in the scene. The thermal sensor can identify the humans based on the variation of temperature in the region. The drawback of the counter using thermal sensors was that if there exist any change in temperature in the environment, the system might get corrupted. In case the temperature of the surroundings is close to the human body temperature, the system fails to respond.

Other people counting systems include the foot step sensor or the sound sensor. These were no used so often as they need very supportive environment and high maintenance. These did not even produce the result regarding the direction of movement of the people which determines the number of people who were entering and leaving that particular area. The most preferable method for counting people is using the image processing algorithms based on the computer-vision- based techniques. Research is still going on in this field and in recent days , the people count system is being widely used in many places including shopping malls, colleges, industries, tourist places, places which are occupied with denser crowds.

Counting systems based on computer-vision provides an effective path for this work. The foremost thing to be done in the computer-vision systems is to segregate the moving object (in this case people) from the static background scene. Through which the process of the foreground extraction can be done.

One of the methods of people detection was the use of the virtual lines for the people count. This involves the counting of the people when the virtual line in the field of view of camera had been crossed by any obstruction. This had a drawback which was because of the usage of the single virtual line that increments the counter when a person crosses the virtual line. In case of people flow after one by one, this method can be accurate. But, in case of the people entering or leaving the place at the same time, there exists ambiguity and the system fails in the proper count of the people.

In order to avoid the ambiguity a raised due to the multiple persons crossing the virtual line at the same time, two virtual lines can be considered which improves the systems performance and the accuracy of the result. By using two virtual lines, one corresponding to the IN count (number of persons entering the place) and the other corresponding to the OUT count (number of persons leaving the place ) are considered.

After the segregation of the foreground objects in the current frame, the task that has to be done was the people detection which can be done using various techniques. By obtaining the foreground extracted features, basic and easiest operations like Hough transform can be done in order to find the circles or disk shaped structures to detect the human. This is not much efficient and can give the result up to 60% along with some false detections.

Another method of segmentation of the people and the background can be done using the background subtraction algorithm which was done through simulating neural networks . This method used an algorithm to avoid occlusions, which was dynamically adjustable based on the situation. The limitation was the speed reduction due to the usage of neural network. This algorithm was applicable to people count in a particular image which did not consider the people tracking.

### III. Proposed Algorithm

The proposed algorithm for people count in a video is implemented in OPEN CV3 using python 3.5.2. The implementation goes on like the process of background subtraction on the incoming frames followed by the blob analysis using which the person can be detected and by using virtual lines, the count of the people entering and leaving a particular area can be evaluated.

The presented people counting algorithm is implemented using blob analysis with the basic theory of background subtraction, which is the foremost step in order to detect the moving objects. In a video generally, background subtraction has to be done to separate the moving objects from the static objects. The blob is to be detected for the foreground extracted objects and based on the blobs detected, the heads of the people can be identified in each frame. The blob detection is possible by considering the blob parameters like circularity, convexity, inertia. The detected blobs return the diameter of the blob along with its centroid. The blob detection is used to identify the presence of human based on the head part.

The blob detected returns the key points corresponding to the Centre and diameter of the blob (x,y) and s respectively. The count can be updated based on the tracking of the blob detected. The separation of the directional movement can be done by using virtual lines. By matching the centroids of the blob with the virtual lines, the IN-Count and OUT-Count can be differentiated.

### IV. Results & Discussion

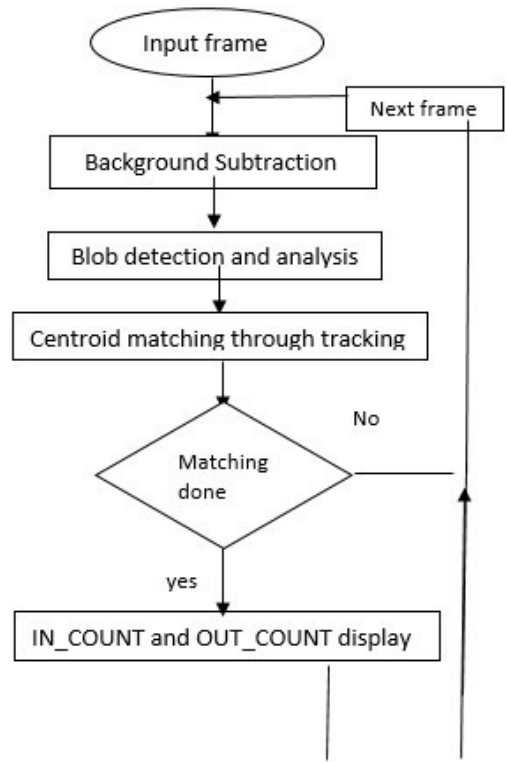

**Figure 1.** Block Diagram Of Proposed Algorithm

The people counting algorithm implemented has tested at the entrance of a laboratory which provides the information regarding the number of people entering and leaving that particular laboratory. A video is captured for about a particular period of time. The captured video is given as input video to the algorithm implemented and the obtained results are compared with the actual number of people entering and leaving the laboratory. The results obtained by the proposed algorithm are as follows:

The evaluation of the algorithm done by the comparison of the IN-COUNT and OUT-COUNT obtained with the exact or the actual IN-COUNT and OUT-COUNT. The evaluation results can be tabulated as follows:

**Table 1.** Tabular Column For In-count and Out-count

| Sn | I | Ou | True | True | Totalcou | Truetotalcou |
|---|---|---|---|---|---|---|

| o | n | t | -in | -out | nt | nt |
|---|---|---|-----|------|-----|-----|
| 1 | 8 | 8 | 8 | 8 | 16 | 16 |
| 2 | 9 | 12 | 9 | 12 | 21 | 21 |
| 3 | 20 | 25 | 20 | 28 | 45 | 48 |

In the consideration of the results obtained using the proposed algorithm, a graph is plotted in comparison to the true count and the resultant count. The results plotted are shown below in a graphical manner:

In accordance to the results obtained, the accuracy of the results can be calculated as :

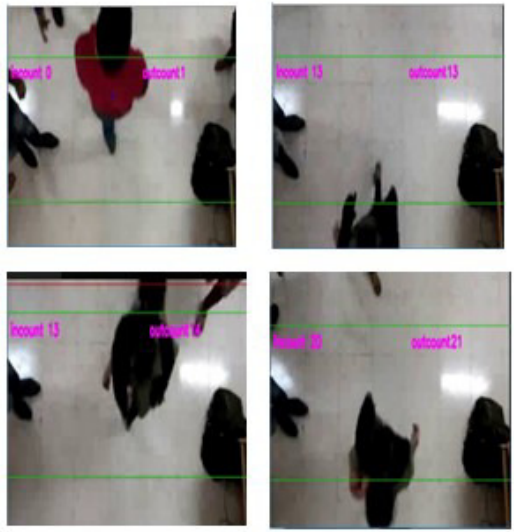

**Figure 2.** In-Count And Out-Count Display

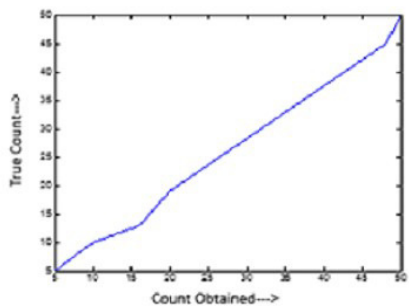

**Figure 3.** Comparison Of True Count with Actual Count

$$In\_accuracy\ \% = \frac{In\_count}{True\_InCount} * 100$$

$$Out\_accuracy\ \% = \frac{Out\_Count}{True\_Out\ Count} * 100$$

$$TC\ accuracy\ \% = \frac{Total\ Count}{True\ Total\ count} * 100$$

**Table 2.** Depicting the accuracy of the proposed algorithm

| Sno | In-accuracy% | Out-accuracy% | TC-accuracy% |
|-----|--------------|---------------|--------------|
| 1 | 100 | 100 | 100 |
| 2 | 100 | 100 | 100 |
| 3 | 96 | 88 | 94 |

The evaluation of the proposed algorithm results in an accuracy of 93% to 94%.

## V. Conclusion

The obtained results reflect the efficiency of the proposed algorithm. The algorithm implemented, results in an accuracy of 93% which can be practically used for the applications of people counting. As the proposed work uses the blob analysis that identifies the person by his/her head, occlusion can be avoided that improves the efficiency of algorithm leading to exact count of the people. In addition to the occlusion problem, the algorithm proposed has overcome the more usage of memory in order to train the datasets. The differentiation of the human from any other creatures can be done by the consideration of certain geometric parameters and the results obtained are accurate and reliable. Most practical applications of the people count systems installed, includes the crowded areas to estimate the number of people and shopping malls etc.

### Acknowledgement

The presented work was supported by the principal of KL University and also the support of management means a lot, who provided allowance to capture live video samples for the completion of the work.